\documentclass{article}
\usepackage{abstract}

\usepackage[T1]{fontenc}       
\usepackage{lmodern}           

\usepackage{amsmath}           
\usepackage{amssymb}           
\usepackage{bm}                
\usepackage{mathtools}         
\usepackage{newtxmath}         

\usepackage{graphicx}          
\usepackage{subfigure}         
\usepackage{xcolor}            
\definecolor{darkblue}{rgb}{0,0.08,0.45} 
\usepackage{wrapfig} 

\usepackage{xurl}              
\usepackage{hyperref}          
\hypersetup{
    breaklinks=true,
    colorlinks=true,
    urlcolor=darkblue,
    linkcolor=darkblue,
    citecolor=darkblue
}

\usepackage{booktabs}          
\usepackage{multirow}          
\usepackage{makecell}          
\usepackage{float} 
\usepackage{natbib} 
\usepackage{lipsum} 
\usepackage{marginnote}
\usepackage{textpos}
\usepackage{tabularx} 
\usepackage{booktabs}
\usepackage{xcolor}
\usepackage{colortbl}
\usepackage{graphicx}

\usepackage{enumerate}         
\usepackage{enumitem}          

\usepackage{mdframed}
\usepackage{tcolorbox}

\usepackage[accepted]{icml2025}

\usepackage{amsthm}            

\usepackage{caption}           
\usepackage[capitalize,noabbrev]{cleveref}

\usepackage{footmisc}
\usepackage{booktabs}
\usepackage{graphicx}
\usepackage{stfloats} 

\usepackage{graphicx}
\usepackage{placeins}
\theoremstyle{plain}

\theoremstyle{definition}

\theoremstyle{remark}

\usepackage[textsize=tiny]{todonotes}

\usepackage{pifont}

\icmltitlerunning{HODDI}

\begin{document}

\twocolumn[
\icmltitle{HODDI: A Dataset of High-Order Drug-Drug Interactions for Computational Pharmacovigilance}



\icmlsetsymbol{equal}{*}

\begin{icmlauthorlist}
\icmlauthor{Zhaoying Wang}{iit}
\icmlauthor{Yingdan Shi}{iit}
\icmlauthor{Xiang Liu}{iit}
\icmlauthor{Can Chen}{unc}
\icmlauthor{Jun Wen}{harv}
\icmlauthor{Ren Wang}{iit}
\end{icmlauthorlist}

\icmlaffiliation{iit}{Illinois Institute of Technology}
\icmlaffiliation{unc}{The University of North Carolina at Chapel Hill}
\icmlaffiliation{harv}{Harvard Medical School}

\icmlcorrespondingauthor{Can Chen}{canc@unc.edu}
\icmlcorrespondingauthor{Jun Wen}{Jun$\_$Wen@hms.harvard.edu}
\icmlcorrespondingauthor{Ren Wang}{rwang74@iit.edu}
\vskip 0.3in
]



\printAffiliationsAndNotice{}  

\newcommand{\authnote}[2]{{\bf \textcolor{blue}{#1}: \em \textcolor{red}{#2}}}
\newcommand{\Ren}[1]{\authnote{Ren}{#1}}
\newcommand{\Yingdan}[1]{\authnote{Yingdan}{#1}}
\newcommand{\Zhaoying}[1]{\authnote{Zhaoying}{#1}}

\newcommand{\nodes}{\mathcal{V}}
\newcommand{\edges}{\mathcal{E}}
\newcommand{\node}{\mathcal{v}}

\newcommand{\trn}{G^{tr}}
\newcommand{\Dclf}{G^{clf}}
\newcommand{\Dwmk}{G^{wmk}}
\newcommand{\val}{G^{val}}
\newcommand{\tst}{G^{test}}

\newcommand{\Xtrn}{\textbf{X}^{tr}}
\newcommand{\Xclf}{\textbf{X}^{clf}}
\newcommand{\Xwmk}{\textbf{X}^{wmk}}
\newcommand{\Xval}{\textbf{X}^{val}}
\newcommand{\Xtst}{\textbf{X}^{test}}
\newcommand{\ytrn}{{\bf y}^{tr}}
\newcommand{\yclf}{{\bf y}^{clf}}
\newcommand{\ytr}{{\bf y}^{clf}}
\newcommand{\ywmk}{{\bf y}^{wmk}}
\newcommand{\yval}{{\bf y}^{val}}
\newcommand{\ytst}{{\bf y}^{test}}
\newcommand{\ytrnhat}{{\bf p}^{tr}}
\newcommand{\yclfhat}{{\bf p}^{clf}}
\newcommand{\ywmkhat}{{\bf p}^{wmk}}
\newcommand{\yvalhat}{{\bf p}^{val}}
\newcommand{\ytsthat}{{\bf p}^{test}}
\newcommand{\edgestrn}{\edges^{tr}}
\newcommand{\edgesclf}{\edges^{clf}}
\newcommand{\edgeswmk}{\edges^{wmk}}
\newcommand{\edgesval}{\edges^{val}}
\newcommand{\edgestst}{\edges^{test}}
\newcommand{\nodestrn}{\nodes^{tr}}
\newcommand{\nodesclf}{\nodes^{clf}}
\newcommand{\nodeswmk}{\nodes^{wmk}}
\newcommand{\nodesval}{\nodes^{val}}
\newcommand{\nodestst}{\nodes^{test}}
\newcommand{\e}{\textbf{e}}
\newcommand{\Ehat}{\hat{\e}}
\newcommand{\Ewmk}{\e^{wmk}}
\newcommand{\idxwmk}{\textbf{idx}}
\newcommand{\numsub}{T}
\newcommand{\Ewmkoneindexed}{\Ewmk_1[\idxwmk]}
\newcommand{\Ewmkiindexed}{\Ewmk_i[\idxwmk]}
\newcommand{\Ewmktindexed}{\Ewmk_{\numsub}[\idxwmk]}
\newcommand{\Ewmkhat}{\Ehat^{wmk}}
\newcommand{\Ewmkhatoneindexed}{\Ewmkhat[\idxwmk]}
\newcommand{\Ewmkhatiindexed}{\Ewmkhat_i[\idxwmk]}
\newcommand{\Ewmkhatkindexed}{\Ewmkhat_k[\idxwmk]}
\newcommand{\Ecdt}{\e^{cdt}}
\newcommand{\Ecdtoneindexed}{\Ecdt_1[\idxwmk]}
\newcommand{\Ecdtiindexed}{\Ecdt_i[\idxwmk]}
\newcommand{\Ecdttindexed}{\Ecdt_{\numsub}[\idxwmk]}
\newcommand{\Ecdthat}{\Ehat^{cdt}}
\newcommand{\Ecdthatoneindexed}{\Ecdthat[\idxwmk]}
\newcommand{\Ecdthatiindexed}{\Ecdthat_i[\idxwmk]}
\newcommand{\Ecdthatkindexed}{\Ecdthat_k[\idxwmk]}
\newcommand{\Lwmkf}{\mathcal{L}_{wmk}}
\newcommand{\Lclff}{\mathcal{L}_{clf}}
\newcommand{\Lwmk}{\mathcal{L}_{wmk}}
\newcommand{\Lclf}{\mathcal{L}_{clf}}
\newcommand{\Lce}{\mathcal{L}_{CE}}
\newcommand{\bigH}{\textit{\textbf{H}}}
\newcommand{\bigI}{\textit{\textbf{I}}}
\newcommand{\bigK}{\textit{\textbf{K}}}
\newcommand{\barK}{\bar{\bigK}}
\newcommand{\bigL}{\textit{\textbf{L}}}
\newcommand{\barL}{\bar{\bigL}}
\newcommand{\tilK}{\tilde{\bigK}}
\newcommand{\tilL}{\tilde{\bigL}}
\newcommand{\HKH}{\bigH\bigK^{(k)}\bigH}
\newcommand{\HLH}{\bigH\bigL\bigH}
\newcommand{\Xuk}{\textbf{x}^{(k)}_u}
\newcommand{\Xvk}{\textbf{x}^{(k)}_v}
\newcommand{\wmk}{\textbf{w}}
\newcommand{\MIcdt}{\text{MI}^{cdt}}
\newcommand{\MItgt}{\text{MI}^{tgt}}
\newcommand{\MIlb}{\text{MI}^{LB}}
\newcommand{\atgt}{\alpha_{tgt}}
\newcommand{\alb}{\alpha_{LB}}
\newcommand{\av}{\alpha_{v}}
\newcommand{\signate}{\sigma_{nat_e}}
\newcommand{\signatp}{\sigma_{nat_p}}
\newcommand{\munate}{\mu_{nat_e}}
\newcommand{\munatp}{\mu_{nat_p}}
\newcommand{\zlb}{z_{LB}}
\newcommand{\zv}{z_{v}}
\newcommand{\nsub}{n_{sub}}
\newcommand{\explFn}{explain}
\newcommand{\BW}[1]{\textcolor{red}{[AW: ~#1]}}
\newcommand{\JD}[1]{\textcolor{blue}{[JD: ~#1]}}
\newcommand{\wmksubgraph}{G^{wmk}}
\newcommand{\wmksubgraphs}{\{\wmksubgraph_1 ,\dots, \wmksubgraph_{\numsub}\}}
\newcommand{\fix}{\marginpar{FIX}}
\newcommand{\new}{\marginpar{NEW}}

\begin{abstract}
Drug-side effect research is vital for understanding adverse reactions arising in complex multi-drug therapies. However, the scarcity of higher-order datasets that capture the combinatorial effects of multiple drugs severely limits progress in this field. Existing resources such as TWOSIDES primarily focus on pairwise interactions. To fill this critical gap, we introduce \textbf{HODDI}, the first \underline{H}igher-\underline{O}rder \underline{D}rug-\underline{D}rug \underline{I}nteraction Dataset, constructed from U.S. Food and Drug Administration (FDA) Adverse Event Reporting System (FAERS) records spanning the past decade, to advance computational pharmacovigilance. 
HODDI contains 109,744 records involving 2,506 unique drugs and 4,569 unique side effects, specifically curated to capture multi-drug interactions and their collective impact on adverse effects. Comprehensive statistical analyses demonstrate HODDI's extensive coverage and robust analytical metrics, making it a valuable resource for studying higher-order drug relationships. Evaluating HODDI with multiple models, we found that simple Multi-Layer Perceptron (MLP) can outperform graph models, while hypergraph models demonstrate superior performance in capturing complex multi-drug interactions, further validating HODDI's effectiveness. 
Our findings highlight the inherent value of higher-order information in drug-side effect prediction and position HODDI as a benchmark dataset for advancing research in pharmacovigilance, drug safety, and personalized medicine. The dataset and codes are available at \url{https://github.com/TIML-Group/HODDI}.
\end{abstract}

\section{Introduction}
\label{intro}

Clinical observations have revealed numerous cases in which drug combinations lead to unexpected adverse effects \cite{tekin2017measuring,tekin2018prevalence}. For instance, the concurrent use of multiple antibiotics with anticoagulants can significantly increase bleeding risks \cite{baillargeon2012concurrent}, while the combination of various antidepressants may elevate the likelihood of serotonin syndrome \cite{quinn2009linezolid}. These scenarios underscore the urgent need to investigate higher-order drug-drug interactions \cite{tekin2018prevalence}, as understanding the intricate relationships between multiple drugs and their side effects is pivotal in pharmacology, forming the foundation for pharmacovigilance and drug safety studies \cite{tatonetti2012data, salas2022use, huang2021moltrans}. By elucidating these interactions, researchers can identify potential adverse effects early in the drug development process \cite{montastruc2006pharmacovigilance, kompa2022artificial}, optimize drug design \cite{lipinski2012experimental}, mitigate polypharmacy risks \cite{lukavcivsin2019emergent}, and advance personalized medicine \cite{molina2017personalized} by helping clinicians tailor treatments to individual patient characteristics \cite{ryu2018deep}. Nevertheless, current research faces two major challenges: the scarcity of datasets capturing complex, high-order drug-drug interactions \cite{zitnik2018modeling, wang2017pubchem}, and the limitations of existing computational methods in effectively modeling these high-order relationships \cite{masumshah2021neural, sachdev2020comprehensive, zakharov2014computational}.


Several existing datasets have been developed for the research of drug-side effect relationships. While these datasets are valuable resources, they have inherent limitations in capturing high-order drug interactions.  SIDER~\cite{kuhn2010side,kuhn2016sider} and OFFSIDES~\cite{kumar2024predicting} focus primarily on single-drug effects, while TWOSIDES~\cite{tatonetti2012data} advances to pairwise drug combinations. Despite being augmented by biological interaction networks  (e.g., HPRD~\cite{keshava2009human}, BioGRID~\cite{stark2006biogrid}, STITCH~\cite{kuhn2007stitch}), these datasets predominantly focus on single-drug or two-drug interactions, facing significant challenges in polypharmacy scenarios due to insufficient coverage of multi-drug combinations~\cite{tekin2017measuring}, inconsistent data quality~\cite{ismail2022fda}, and the exponential growth of possible drug combinations relative to available data~\cite{tekin2018prevalence}.

Various computational methods have been developed for drug-side effect relationship prediction. Traditional matrix factorization approaches, such as Non-negative Matrix Factorization (NMF) \cite{lee1999learning} and Probabilistic Matrix Factorization (PMF) \cite{jain2023graph}, face challenges in computational complexity due to high-dimensional matrix operations \cite{jain2023graph}, while traditional machine learning methods struggled with modeling complex non-linear relationships in drug-adverse effect interactions \cite{lee1999learning}. Deep learning frameworks, such as the combination of Convolutional Neural Networks (CNNs) and Long Short-Term Memory networks (LSTMs), while improving spatial and temporal feature extraction capabilities, exhibit limitations in molecular structure representation \cite{karim2019drug, peng2024effective, xu2018leveraging}. Graph-based methods like Decagon and Graph Convolutional Network (GCN) demonstrated promise by leveraging protein interaction networks \cite{zitnik2018modeling}, but were primarily restricted to modeling pairwise drug interactions \cite{zitnik2018modeling, zhang2023emerging}. Due to the lack of specialized datasets capturing higher-order drug-adverse effect relationships, existing methods largely rely on datasets containing only pairwise drug interactions, fundamentally limiting their capability to model complex polypharmacy scenarios \cite{tekin2017measuring, tekin2018prevalence}.

To address this gap, we introduce the \textbf{H}igher-\textbf{O}rder \textbf{D}rug-\textbf{D}rug \textbf{I}nteraction (\textbf{HODDI}) dataset, the first resource specifically designed to capture higher-order drug-drug interactions and their associated side effects. HODDI is constructed from the FAERS database \cite{fda_faers, faers_technical} and consists of 109,744 records spanning the past 10 years, covering 2,506 unique drugs and 4,569 distinct side effects. Through rigorous data cleaning and conditional filtering, we focused on cases of co-administered drugs, enabling the study of their combinational impacts on adverse effects. Comprehensive statistical analyses characterize the dataset’s key properties, highlighting its potential for advancing polypharmacy research.

To assess HODDI's utility, we created evaluation subsets and tested its performance using multiple models, with detailed methods described in Sections~\ref{sec:subsets} and \ref{Benchmark Methods}. Our experiments reveal two key findings about leveraging higher-order information in drug-side effect prediction: \ding{202} Even simple architectures like Multi-Layer Perceptron (MLP) can achieve strong performance when utilizing higher-order features from our dataset, sometimes outperforming more sophisticated models like Graph Attention Network (GAT). This suggests the inherent value of higher-order drug interaction data; \ding{203} Models that explicitly incorporate hypergraph structures (e.g., HyGNN \cite{saifuddin2023hygnn}) can further enhance prediction accuracy by better capturing complex multi-drug relationships.

We summarize our contributions as follows:
\begin{itemize}[leftmargin=*]
\vspace{-2mm}
    \item We constructed \textbf{HODDI}, the first dataset focusing on higher-order drug-drug interactions from the FAERS database.
    \item We comprehensively analyzed HODDI's statistical characteristics across multiple time scales.
    \item Our experiments highlight the importance of higher-order information in drug-side effect prediction and demonstrate the effectiveness of the proposed hypergraph model.

\end{itemize}
\section{Related Work}
\label{sec:related}
\paragraph{Existing Datasets.} Several specialized medical datasets have been developed for studying drug-side effect relationships. SIDER~\cite{kuhn2010side,kuhn2016sider} contains drug-side effect associations extracted from drug labels and public documents, serving as a foundational dataset of single drug documentation. Building upon this, OFFSIDES~\cite{kumar2024predicting} incorporates data from clinical trials, offering more rigorous evidence. TWOSIDES~\cite{tatonetti2012data} further advances the data complexity by providing comprehensive information about drug combinations and their associated side effects derived from FAERS~\cite{fda_faers,faers_technical}, capturing real-world multi-drug interaction cases. These datasets are further augmented by protein-protein interaction networks from databases like HPRD~\cite{keshava2009human} and BioGRID~\cite{stark2006biogrid}, and drug-protein target interactions from STITCH~\cite{kuhn2007stitch}, enabling researchers to construct multi-modal graphs for side effect prediction in both single-drug and combination therapy scenarios \cite{zitnik2018modeling, wang2009pubchem, ryu2018deep, vilar2017role}. While these datasets have achieved some success in characterizing drug-side effect relationships, they face significant limitations in capturing higher-order interactions in polypharmacy scenarios. The challenges stem from three main aspects: \ding{202} Insufficient coverage of drug combinations, which limits our understanding of complex drug interactions; \ding{203} Inconsistent data quality across different sources~\cite{ismail2022fda}, which affects the reliability of predictions; and \ding{204} Data sparsity issues in polypharmacy settings, where the number of possible drug combinations grows exponentially while available data remains limited~\cite{tekin2017measuring,tekin2018prevalence}.

\paragraph{Existing Methods.} 
Early research primarily relied on traditional machine learning methods, including Naive Bayes \cite{jansen2003bayesian, burger2008accurate}, Support Vector Machines (SVMs) \cite{ferdousi2017computational}, and Network Analysis \cite{ye2014construction}, which were used to analyze chemical structures \cite{staszak2022machine}, molecular fingerprints \cite{rogers2010extended}, and protein interaction networks \cite{chowdhary2009bayesian, bradford2006insights}. Subsequently, matrix factorization approaches such as NMF \cite{lee1999learning} and PMF \cite{jain2023graph} were introduced to decompose drug interaction matrices, though challenged by computational complexity and interpretability. With the rise of deep learning, CNNs and LSTMs were combined for the extraction of spatial features from molecular structures and the capture of temporal dependencies in drug interactions, effectively modeling complex drug-drug interaction relationships, particularly in polypharmacy scenarios \cite{xu2018leveraging, karim2019drug, wen2023multimodal, peng2024effective}. Due to the inherent advantages on leveraging the established biomedical relations, graph-based methods have been predominately investigated. For instance, the Decagon model \cite{zitnik2018modeling} leverages graphs to leverage both the drug-drug and protein-protein interactions to connect drug to side effects, and Graph Attention Networks (GAT) \cite{mohamedtrivec} were leveraged attention mechanisms to improve performance. Despite these advancements, challenges remain due to the scarcity of specialized datasets capturing higher-order drug-drug interaction relationships \cite{lukavcivsin2019emergent}, particularly in polypharmacy scenarios where complex multi-drug interactions need to be modeled \cite{fatima2022comprehensive}. While researchers have achieved some success using benchmark datasets like DrugBank \cite{wishart2018drugbank} and TWOSIDES \cite{tatonetti2012data, zhang2023emerging}, the current model performance remains limited by the lack of datasets containing higher-order drug-adverse effect relationships.



\section{HODDI Dataset Construction}
\label{dataset_construction}

We utilized the FAERS database~\cite{fda_faers,faers_technical} as our primary data source, which contains comprehensive adverse event reports, medication error reports, and product quality complaints submitted to the U.S. Food and Drug Administration (FDA) by healthcare professionals, consumers, and manufacturers. We downloaded the quarterly datasets spanning from 2014Q3 to 2024Q3, covering the latest 41 quarters in total.\footnote{Q3 denotes the third quarter of the year.} Each quarterly dataset is a compressed zip file containing individual raw XML files.

\paragraph{Key Components Extraction.}
After decompressing the downloaded zip files and parsing the XML data, we extracted key components from each record in the FAERS database, including the report ID, the side effect names, and the standardized drug names with their drug roles\protect\footnotemark\kern0.2em.

\begin{table}[h]
    \centering
    \caption{An example of the extracted components from a FAERS record: Three medications—Hydroxychloroquine, Prednisone, and Terbinafine—were administered simultaneously, all classified under Drug Role 3 (concomitant medications).}
    \label{extracted_component}
    \resizebox{0.45\textwidth}{!}{
    \begin{tabular}{ccc}
    \toprule
    \rowcolor{gray!20}
    \textbf{Report ID} & \textbf{Side Effect} & \textbf{Drug (Role)} \\
    \midrule
    24135951 & \begin{tabular}[c]{@{}c@{}}Psoriasis\\Drug interaction\end{tabular} & \begin{tabular}[c]{@{}c@{}}Hydroxychloroquine (3)\\Prednisone (3)\\Terbinafine (3)\end{tabular} \\
    \bottomrule
    \end{tabular}}
\end{table}

\begin{@twocolumnfalse}
\footnotetext{Drug roles are categorized into three types in the FAERS database: \textbf{a) Drug Role 1:} primary suspect drugs that are primarily suspected of causing the adverse event; \textbf{b) Drug Role 2:} secondary suspect drugs that may have contributed to the event; \textbf{c) Drug Role 3:} concomitant medications that were being taken at the same time of the event.}
\end{@twocolumnfalse}

Table~\ref{extracted_component} illustrates an example of a simultaneous drug administration during an adverse event period. According to the clinical report (ID: 24135951), three concomitant medications (Hydroxychloroquine, Prednisone, and Terbinafine), classified as Drug Role 3, were administered concurrently during the treatment. This combination of drug therapy represents a complex pharmacological intervention with potentially risky drug-drug interactions.

\paragraph{Conditional Filtering.}

After extraction of key components from each record in the FAERS database, the resulting records from the latest 41 quarters (2014Q4-2024Q3) were filtered based on the following three conditions to capture the potential suspect drug interactions:

\textbf{a) Condition 1:} At least two concomitant drugs \textit{(Drug Role 3)} were reported in a record, with no primary suspect drugs \textit{(Drug Role 1)} or secondary suspect drugs \textit{(Drug Role 2)}; 

\textbf{b) Condition 2:} At least one primary suspect drug \textit{(Drug Role 1)} combined with at least one concomitant drug \textit{(Drug Role 3)}, with no secondary suspect drugs \textit{(Drug Role 2)}; 

\textbf{c) Condition 3: }Concurrent presence of all three drug roles - primary suspect drug \textit{(Drug Role 1)}, secondary suspect drug \textit{(Drug Role 2)}, and concomitant drug \textit{(Drug Role 3)}. 

These conditions enable comprehensive identification of potential adverse drug interactions, including cases where concomitant medications may contribute to adverse events.

\paragraph{Side Effects Processing.}

Side effect names were first extracted from the original FAERS records and then encoded into 768-dimensional embedding vectors using Self-aligning pretrained Bidirectional Encoder Representations from Transformers (SapBERT)~\cite{liu-etal-2021-self}, a model pre-trained on PubMed texts for biomedical entity normalization~\cite{lim2022sapbert}. 
The mapping between the standardized adverse event terminology (MedDRA terms) and their corresponding Unified Medical Language System (UMLS) Concept Unique Identifiers (CUIs) were obtained from the UMLS Metathesaurus~\cite{umls}.
The cosine similarities were calculated between all embedding pairs of the side effect names from the original FAERS records and the standardized MedDRA terms, with their embeddings generated using SapBERT.
For each side effect name in the records, its recommended CUI was determined by identifying the standardized MedDRA term with the highest cosine similarity to its embedding.
The cosine similarity thresholds of 0.8 and 0.9 were used to stratify the side effects based on their confidence levels, with the higher threshold value indicating the higher confidence of their classifications as recommended CUIs. 
The recommended CUIs served as the labels of side effect names, while the 768-dimensional embedding vectors of the corresponding standardized MedDRA terms served as the feature representations across all models.

\paragraph{Drug Names Processing.}

The standardized drug names were extracted from the FAERS records and normalized by converting all standardized drug names to uppercase, removing salt form suffixes (e.g., hydrochloride/HCL), and handling compound drug names.
Then the normalized drug names were mapped to their corresponding DrugBank IDs using DrugBank's comprehensive database (DrugBankFullDatabase.xml)~\cite{wishart2006drugbank}, which contains the detailed drug information including synonyms and alternative names.

\paragraph{Positive Sample Construction.}

In the context of polypharmacy, positive samples refer to drug combinations that have been confirmed to cause specific side effects in real-world medical records. For HODDI dataset construction, positive samples were constructed from records with high-confidence side effects with high cosine similarities above 0.9 for side effect embeddings generated by SapBERT. To eliminate data redundancy, we removed duplicate records and checked for superset relationships. For each record with a higher number of co-administered drugs, we examined whether its drug combination and associated side effects completely overlapped with those in records containing fewer drugs. If such complete overlap was found, we removed the superset record to maintain the most parsimonious representation of drug-side effect associations across the dataset spanning 41 quarters (2014Q3-2024Q3). This reduction strategy ensures each unique drug combination is represented only by its minimal complete set, preserving the most fundamental drug interactions while eliminating redundant higher-order drug combinations, improving computational efficiency and reducing noises in the HODDI dataset.

\paragraph{Negative Sample Construction.}
Negative samples represent drug combinations that have not been reported to cause the specific adverse events in clinical records or are unlikely to trigger such adverse events based on their known pharmacological properties.
In this work, negative samples were artificially constructed cases by replacing both a drug and a side effect from positive samples with random alternatives, based on the assumption that it is extremely unlikely for a randomly assembled drug combination to cause a specific side effect while ensuring the generated samples don't exist among positive samples. An equal number of positive and negative samples were generated, which helps maintain the class balance in the HODDI dataset and prevents model bias during training, ensuring that the model would not favor predicting one class over another, and improving its generalization capability and prediction accuracy.
The obtained positive and negative samples were partitioned chronologically into 41 quarterly intervals from 2014Q3 to 2024Q3, resulting in 41 pairs of positive and negative sample files (82 CSV files in total).


\begin{table}[t] 
   \centering
    \caption{HODDI dataset statistics, including the number of records (\#Record), unique standardized drug names per record in positive/negative samples (\#Drug+/-), and unique CUIs in positive/negative samples (\#CUI+/-).}
   \setlength{\tabcolsep}{4pt}
   \centering
   \resizebox{0.48\textwidth}{!}{
   \begin{tabular}{@{}cccccc@{}}
       \toprule
       \rowcolor{gray!20}
       \textbf{\large Time} & \textbf{\large \#Record} & \textbf{\large \#Drug+} & \textbf{\large \#Drug-} & \textbf{\large \#CUI+} & \textbf{\large \#CUI-} \\ 
       \midrule
       \textbf{2014Q3-2024Q3} & 109,744 & 2,506 & 12,293 & 3,950 & 4,581 \\
       \bottomrule
   \end{tabular}
   }
   \vspace{-4mm} 

   \label{dataset_statistics_41quarters}
   \vspace{-2mm}
\end{table}

\vspace{-2mm}
\section{Data Analysis}
\label{data_analysis}

\paragraph{Dataset Structure.}
The HODDI dataset comprises two distinct sets of CSV files: The first set contains the merged positive and negative samples consolidated into two CSV files (\textit{pos.csv} and \textit{neg.csv}). The second set encompasses quarterly data spanning from 2014Q3 to 2024Q3, containing 82 CSV files (41 quarters \(\times\) 2 files per quarter). For each quarter \(Q\), there are two corresponding files denoted as \(Q\_pos.csv\) and \(Q\_neg.csv\), where \(Q\) represents the quarter identifier (e.g., 2014Q3), and the suffixes \(pos\) and \(neg\) indicate positive and negative samples, respectively.
This dataset structure follows the established pharmacovigilance data processing methods~\cite{sakaeda2013data,poluzzi2012data}. Each CSV file contains five essential columns that capture key information about drug-side effect reports: report identifier, recommended CUI, DrugBank ID, hyperedge labels (1 for positive samples, -1 for negative samples), and temporal information (year and quarter of each record), shown in Table~\ref{hgnn_dataset_structure}. 
The HODDI dataset structure facilitates the analysis of higher-order relationships between drug combinations and their adverse effects, enabling comprehensive data mining of multi-drug interaction pattern.

\paragraph{Dataset Composition.}
As shown in Table~\ref{dataset_statistics_41quarters}, the HODDI dataset spanning from 2014Q3 to 2024Q3 comprises 109,744 drug-side effect association records (balanced between positive and negative samples). In the complete dataset, the positive samples contain 2,506 unique standardized drug names and 3,950 unique CUIs, while the negative samples include 12,293 unique drugs and 4,581 unique CUIs. The increased number of unique entities in the negative samples is attributed to the resampling process during negative sample generation~\cite{ismail2022fda}.


\begin{figure*}[t]
  \centering
  \includegraphics[trim=1 150 2 70, clip, width=\textwidth]{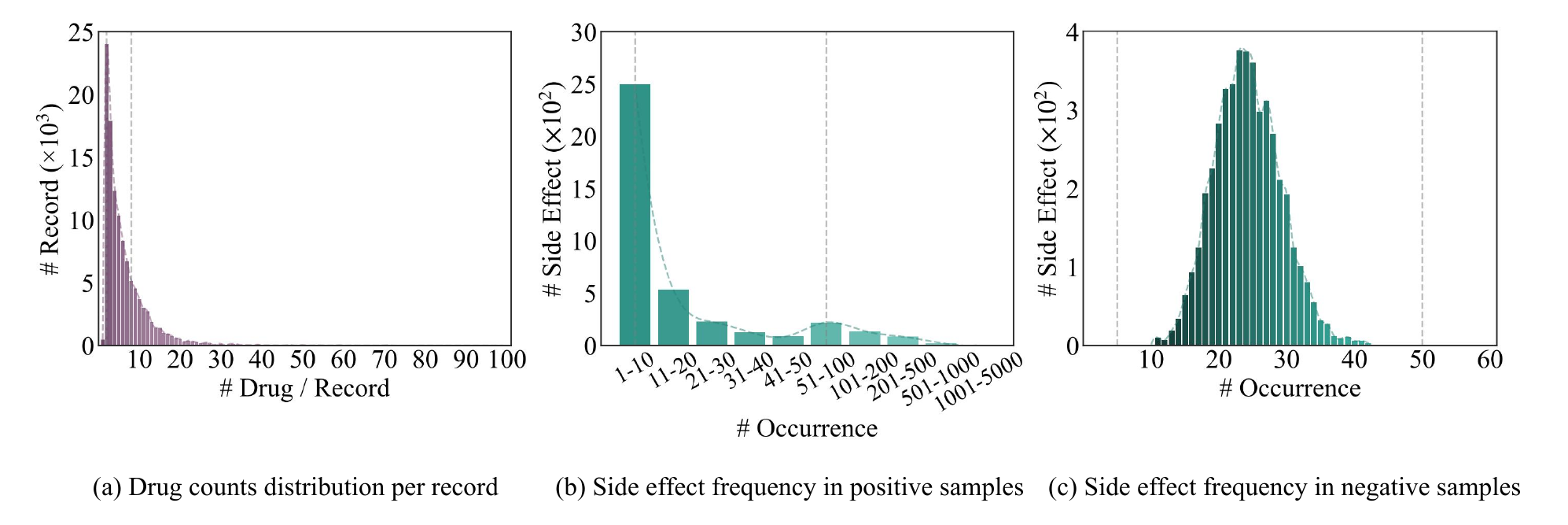}
  \vspace{-8mm}
  \caption{Distribution analysis of medication records and adverse event frequencies in the HODDI dataset (2014Q3-2024Q3). (a) Distribution of drug counts per record (\#Drug/Record), showing a concentration between 2-8 drugs per record; (b) Frequency distribution of adverse event occurrences (\#Occurrence) in positive samples, with most events occurring 1-50 times; (c) Distribution of adverse event occurrences (\#Occurrence) in negative samples, displaying a normal-like distribution centered around 20-30 occurrences. The vertical dashed lines in (a) and (b) mark the intervals of 2-8 \#Drug/Record and 5-50 \#Occurrence, respectively. These intervals were selected as the filtering criteria for our evaluation set due to their high record counts and the most representative higher-order relationships.}
  \label{fig:overall_statistics_drug_se}
\end{figure*}

\paragraph{Drug and Adverse Event Distribution.}
We performed a distribution analysis of the medication records and the adverse event frequencies in the HODDI dataset (2014Q3-2024Q3), with results shown in Figure~\ref{fig:overall_statistics_drug_se}.
The drug count per record exhibits identical distribution for positive and negative samples, whereas the side effect occurrences show distinct distributions between positive and negative samples, as contrasted in Figure \ref{fig:overall_statistics_drug_se}(b) and (c).
This distributional distinction arises from our negative sample generation procedure, which employs random resampling of side effects for negative sample generation.
The drug counts per record in both positive and negative samples show left-skewed distributions with a typical long tail, as shown in Figure \ref{fig:overall_statistics_drug_se}(a). Most records (82\%) contain 2-10 drugs, with very few records (< 1\%) containing more than 30 drugs (maximum: 100).
For side effect occurrences, positive samples show a long-tail distribution spanning from 1 to 5000, with the majority (63.3\%) occurring 1-10 times, as shown in Figure \ref{fig:overall_statistics_drug_se}(b). In contrast, negative samples have a more concentrated and near-normal distribution, with 68.7\% of side effects occurring for 21-30 times within a narrower range of 11-50, as shown in Figure \ref{fig:overall_statistics_drug_se}(c).
Stratified statistics of the medication records and the adverse event frequencies are detailed in Appendix~\ref{Appendix_A1}.

\begin{figure}[h]
\setlength{\abovecaptionskip}{0pt}
\setlength{\belowcaptionskip}{-19pt}
    \centering
    \includegraphics[width=0.9\linewidth, trim=8 5 5 5, clip]{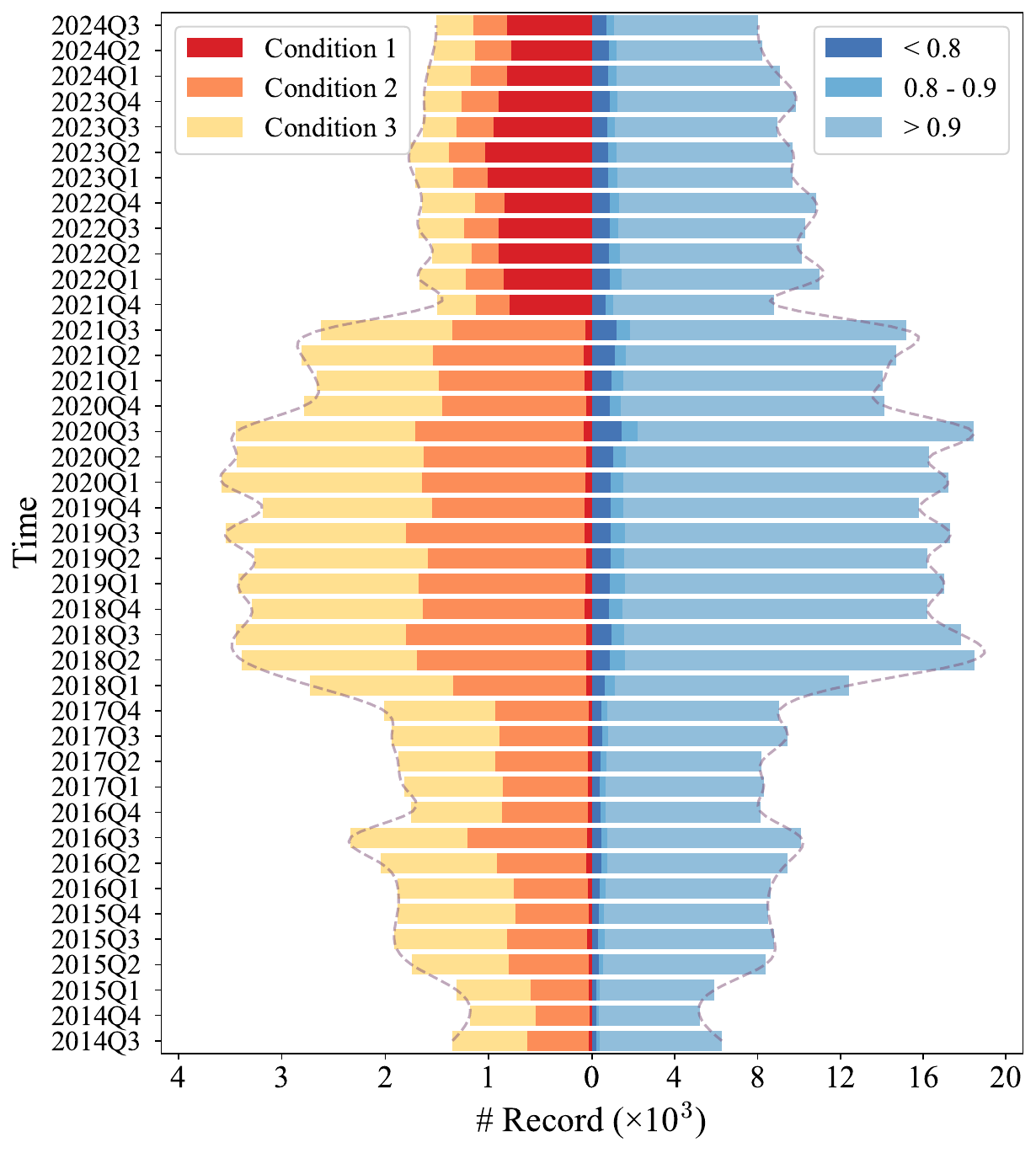}
    
    \vspace{2mm}
    \caption{Temporal trends of the record number of drug conditions (1-3) and cosine similarity ranges (< 0.8, 0.8-0.9, > 0.9) in the HODDI dataset from 2014Q3 to 2024Q3. The horizontal axis shows the number of records (× 10³), with drug conditions displayed on the left and cosine similarity ranges on the right. The vertical axis represents the quarterly time periods of medication records ranging from 2014Q3 to 2024Q3. The stacked bars demonstrate the distribution of records across different categories over quarterly periods.}
    \label{quarterly_trend}
\end{figure}

\paragraph{Drug Interaction Trends.}
Monitoring the temporal trends in suspected drug interactions and validating adverse effect recommendations is essential for pharmacovigilance, enabling tracking of reporting patterns, identification of safety signals, and evaluation of standardization methodologies~\cite{ji2016functional, noren2010temporal}.
To gain these insights, we conducted a longitudinal analysis of the HODDI dataset from 2014Q3 to 2024Q3, examining temporal trends in drug interaction records and the reliability of side effect ULMS CUI recommendations.
Figure~\ref{quarterly_trend} illustrates the quarterly distribution of records across three drug role conditions and side effect cosine similarity ranges between the actual and recommended standardized side effect embeddings.
The raw data is presented in Table~\ref{combined_stats_all_quarters}.
Over the 41 quarters analyzed, the HODDI dataset contains a total of 92,064 drug interaction records, with a mean of 2,246 records per quarter with a standard deviation of 1,412. Condition 3 was the most prevalent drug condition overall, accounting for 46\% of total records, followed by Condition 2 (41\%) and Condition 1 (13\%). However, the relative proportions of the conditions shifted substantially in late 2021.
Prior to 2021Q4, Conditions 2 and 3 were dominant, each averaging approximately 1,500 records per quarter. Starting in 2021Q4, Condition 1 became the most prevalent, consistently recording between 800-1,000 cases per quarter, while Conditions 2 and 3 decreased to around 300-400 records quarterly. This marked shift suggests evolving patterns in drug interaction reporting over time.
The cosine similarity analysis revealed generally high reliability in side effect recommendations, with 90\% of the 470,144 total CUI matches having similarities greater than 0.9. However, we also note that the proportion of low-confidence matches (similarity < 0.8) gradually increased from around 3-4\% in early periods to 8-10\% in recent quarters. This trend points to growing complexity in standardizing side effect terminology.
Our findings highlight the typical temporal trends in the HODDI dataset, with notable changes in the distribution of drug role conditions and a slight decline in recommendation reliability over the study period. Further research is warranted to understand the factors driving these shifts and their implications for drug interaction surveillance.


\paragraph{Drug Distribution Variability Across Quarters.}
We analyzed the drug distribution patterns across 41 quarters (2014Q1-2024Q1). Taking 2024Q3 as an illustrative example, we observed consistent patterns in drug role distributions, as shown in Table~\ref{example_2024Q3_statistics}. Typically, Role 1 medications show the highest presence (mean=1.70 in 2024Q3) and consistent usage patterns (25th-75th percentile: 1-1), while Role 2 exhibits moderate usage (mean=1.51), and Role 3 appears rarely (mean=0.02). The maximum number of drugs per record varies between quarters but consistently shows Role 2 having the highest counts, followed by Role 1, with Role 3 showing notably fewer drugs. While specific values fluctuate across quarters, the general pattern remains: most records contain few drugs (median typically around 1.0) with rare cases of extensive polypharmacy, resulting in a consistent left-skewed distribution pattern~\cite{tekin2018prevalence}.

\renewcommand{\arraystretch}{1.1}
\begin{table}[h]
    \centering
    \caption{An example of quarterly statistics of drug count per record in 2024Q3 HODDI dataset, showing the number of co-administered drug count per record, where Role 1/2/3 indicate different drug roles, and Total represents the sum of drugs across all drug roles (1, 2, and 3) within a single record. This table illustrates the distribution of drugs across different roles and the total drug count within each record.}
    \resizebox{0.4\textwidth}{!}{
    \begin{tabular}{@{}ccccc@{}}
        \toprule
        \rowcolor{gray!20}
        \textbf{\large Metric} & \textbf{\large Role 1} & \textbf{\large Role 2} & \textbf{\large Role 3} & \textbf{\large Total} \\ 
        \midrule
        \textbf{Maximum} & 187 & 200 & 77 & 203 \\
        \textbf{Minimum} & 0 & 0 & 0 & 1 \\
        \textbf{Mean} & 1.70 & 1.51 & 0.02 & 3.23 \\
        \textbf{Median} & 1 & 0 & 0 & 1 \\
        \textbf{25\textsuperscript{th} Percentile} & 1 & 0 & 0 & 1 \\
        \textbf{75\textsuperscript{th} Percentile} & 1 & 1 & 0 & 3 \\
        \bottomrule
    \end{tabular}}
    \label{example_2024Q3_statistics}
\end{table}
\renewcommand{\arraystretch}{1}

\section{Evaluation Subsets Generation}\label{sec:subsets}

To facilitate efficient model evaluation, we constructed three evaluation subsets from the HODDI dataset using the following criteria (excluding low-frequency cases), with their statistical details summarized in Table ~\ref{dataset_statistics_for_3_evaluation_subsets}:

\begin{itemize}[leftmargin=*]
\vspace{-2mm}
    \item \textbf{Evaluation Subset 1:} 21,503 sample pairs where each record contains 2-8 drugs, and each side effect appears for 5-50 times;

    \item \textbf{Evaluation Subset 2:} 24,777 sample pairs where each record contains 2-16 drugs, and each side effect appears for 5-50 times;

    \item \textbf{Evaluation Subset 3:} 38,430 sample pairs where each record contains 2-16 drugs, and each side effect appears for 5-100 times.
\end{itemize}
\vspace{-2mm}
For simplicity, we selected Evaluation Subset 1 as our {\textbf{evaluation set}} for HODDI dataset utility evaluation across graph- and non-graph-based models. One example record from the evaluation set is illustrated in Table~\ref{tab:dataset_example}.
This table presents a detailed example of a single record from the evaluation set, demonstrating the structured format used for capturing adverse drug event information, including unique identifiers for reports and drugs, temporal information, and the binary classification label for potential drug interactions.



\renewcommand{\arraystretch}{1.5} 
\begin{table}[h]
    \centering
    \caption{An example record from evaluation set, showing key attributes including the report identifier (report\_id), the high-confidence UMLS CUI with its cosine similarity above 0.9 (SE\_above\_0.9), the standardized DrugBank identifiers (DrugBankID), the hyperedge label (hyperedge\_label), and the temporal information (time).}
    \label{tab:dataset_example}  
    \resizebox{0.48\textwidth}{!}{
    \begin{tabular}{>{\cellcolor{gray!10}\centering\bfseries}p{1.3cm} >{\centering}p{1.3cm} >{\centering}p{1.3cm} >{\centering}p{1.5cm} >{\centering}p{3cm} >{\centering\arraybackslash}p{0.8cm}}
    \toprule
    \textbf{Attribute} & \textbf{report\_id} & \textbf{SE > 0.9} & \textbf{DrugBankID} & \textbf{hyperedge\_label} & \textbf{time} \\
    \midrule
    \textbf{Value} & 10367822 & C0438167 & \makecell{DB00582 \\ DB08901} & 1 & 2014Q3 \\
    \bottomrule
    \end{tabular}}
\end{table}
\renewcommand{\arraystretch}{1}

\begin{table}[t] 
   \centering
    \caption{HODDI dataset statistics across three evaluation subsets, showing the number of records (\#Records), unique standardized drug names per record in positive/negative samples (\#Drugs+/-), and unique CUIs in positive/negative samples (\#CUIs+/-) }
   \setlength{\tabcolsep}{4pt}
   \centering
   \resizebox{0.47\textwidth}{!}{
   \begin{tabular}{@{}cccccc@{}}
       \toprule
       \rowcolor{gray!20}
       \textbf{\large Time} & \textbf{\large \#Records} & \textbf{\large \#Drugs+} & \textbf{\large \#Drugs-} & \textbf{\large \#CUIs+} & \textbf{\large \#CUIs-} \\ 
       \midrule
       \textbf{Evaluation Subset 1} & 21,503 & 1,779 & 10,460 & 1,420 & 4,540 \\
       \textbf{Evaluation Subset 2} & 24,777 & 2,050 & 10,917 & 1,580 & 4,566 \\
       \textbf{Evaluation Subset 3} & 38,430 & 2,156 & 11,835 & 1,778 & 4,579 \\
       \bottomrule
   \end{tabular}
   }
   \vspace{2mm} 

   \label{dataset_statistics_for_3_evaluation_subsets}
\end{table}

\paragraph{Evaluation Set Conversion.}

To evaluate the HODDI dataset's  applicability to GNN-based models, we implemented the clique expansion method for the evaluation set conversion, due to its superior ability to capture the drug interaction patterns, despite higher computational demands \cite{klamt2009hypergraphs, zhou2006learning}. During evaluation set conversion, each hyperedge was converted into a complete subgraph where every pair of nodes within the hyperedge is connected by an edge, inheriting the hyperedge's properties (report ID, side effect names with recommended ULMS CUIs, and the record time). 
An example record from the \textbf{converted evaluation set} is illustrated in Table~\ref{tab:dataset_example2}.
This example record illustrates the conversion outcome where two drugs (identified as DB00582 and DB08901) form a binary interaction edge, documented in 2014Q3 with a report ID of 10367822. This record exemplifies the preservation of essential hyperedge attributes during their transformation into pairwise drug connections.
A comparative analysis of Tables~\ref{tab:dataset_example} and ~\ref{tab:dataset_example2} demonstrates the transformation process from the original evaluation set to its converted form, revealing that the evaluation set conversion reduces data complexity while sacrificing higher-order drug interaction information.

\begin{table}[h]
    \centering
    \caption{An example record from the converted evaluation set, showing key attributes including the report identifier (report\_id), the connected drugs in graph (source and target), edge label (edge\_label), and the temporal information (time).}
    \label{tab:dataset_example2}  
    \resizebox{0.48\textwidth}{!}{
    \begin{tabular}{>{\cellcolor{gray!10}\centering\bfseries}p{1.5cm} >{\centering}p{1.5cm} >{\centering}p{1.5cm} >{\centering}p{1.5cm} >{\centering}p{1.5cm} >{\centering\arraybackslash}p{1.5cm}}
    \toprule
    \textbf{Attribute} & \textbf{report\_id} & \textbf{source} & \textbf{target} & \textbf{edge\_label} & \textbf{time} \\
    \midrule
    \textbf{Value} & 10367822 & DB00582 & DB08901 & 1 & 2014Q3 \\
    \bottomrule
    \end{tabular}}
\end{table}

\vspace{2mm}

\section{Benchmark Methods}
\label{Benchmark Methods}

To validate the applicability and generalizability of the HODDI dataset, we tested the evaluation set (comprising 21,503 positive and negative samples each) across different representative data-driven deep-learning architectures. Three types of models were employed for this validation:

\begin{enumerate}

    \item \textbf{Multi-Layer Perceptron (MLP)}. As a fundamental neural network architecture, MLP serves as an important baseline for comparison, typical of its simple structure capable of processing flattened input features and easy implementation.
    
    \item \textbf{Traditional Graph Neural Networks (GCN, GAT)}. While limited to pairwise relationships, these models provide effective baseline performance for capturing drug-drug interactions. GCN leverages spectral graph convolutions to aggregate neighborhood information \cite{kipf2016semi}, while GAT employs attention mechanisms to weigh the importance of different node connections \cite{velivckovic2017graph}.

    \item \textbf{Hypergraph Architectures.} The hypergraph architectures excel at capturing higher-order relationships in drug and side effect interactions through the hypergraph structure \cite{10163497}. In this work, we first leverage the HyGNN model, which is a deep learning framework designed to model complex, high-order relationships in data \cite{saifuddin2023hygnn}. In addition, we introduce the \underline{H}yper\underline{g}raph \underline{N}eural \underline{N}etwork with \underline{S}MILES Embedding Features and \underline{A}ttention Mechanism (HGNN-SA), a novel architecture specifically crafted to capture high-order relationships in chemical data efficiently. By combining hypergraph convolution, multi-head attention mechanisms, and molecular embeddings derived from SMILES representations, our approach significantly enhances feature representation and boosts predictive accuracy.

\end{enumerate}

This diverse model selection enables a comprehensive evaluation of how different architectures handle the complex, higher-order relationships present in drug-drug interactions in the HODDI dataset.

\paragraph{Feature Construction.}
For \textit{the features of drugs}, we encoded the SMILES strings of drugs into 768-dimensional embedding vectors using a pretrained SMILES2Vec model \cite{goh2017smiles2vec}. 
Drugs with missing SMILES were excluded from our analysis to ensure data quality and consistency in molecular representation.
For \textit{the features of side effects}, we utilized the SapBERT pretrained model \cite{lim2022sapbert} to encode side effect descriptions into vectors of the same dimensionality.

\paragraph{Data Structures for Benchmarks.}
For \textit{multi-layer perceptron model}, the average drug feature is concatenated with the feature vector of the corresponding side effect of each record. Each concatenated vector with 1536 dimensions forms the input data for the MLP model.
For \textit{traditional graph neural networks}, we constructed heterogeneous graphs consisting of two types of edges: drug-drug interaction edges between drug nodes, and drug-side effect causal relationship edges connecting drug nodes to side effect nodes.
For \textit{hypergraph neural networks}, instead of building pairwise edges, we constructed hyperedges where each hyperedge connects multiple drug nodes that are associated with the same side effect, effectively capturing the drug co-occurrence relationships. Each hyperedge is labeled with its corresponding side effect, maintaining the semantic connection between drug combinations and their side effect.


\section{Results and Analysis}\label{sec:results}

\begin{table}[h]
    \centering
    \caption{Model performance metrics on the evaluation set from the HODDI dataset. Values in parentheses denote the standard deviations. \textbf{Bold} represents the best value in each column. Even basic architectures such as MLP can deliver robust performance when leveraging higher-order features from our dataset, surpassing more complex models like GAT. The hypergraph architectures, such as HyGNN \cite{saifuddin2023hygnn} and the HGNN-SA we designed, take this a step further by enhancing prediction accuracy through its ability to better model intricate multi-drug relationships.}

    \resizebox{0.48\textwidth}{!}{ 
    \begin{tabular}{@{}ccccc@{}}
        \toprule
        \rowcolor{gray!20}
        \textbf{\large Model} & \textbf{\large Precision} & \textbf{\large F1} & \textbf{\large AUC} & \textbf{\large PRAUC} \\
        \midrule
        \bf{HGNN-SA} & \bf{0.906} \textcolor{blue}{(0.002)} & \bf{0.933} \textcolor{blue}{(0.001)} & \bf{0.957} \textcolor{blue}{(0.003)} & \bf{0.939} \textcolor{blue}{(0.008)} \\
        HyGNN & 0.903 \textcolor{blue}{(0.004)} & 0.932 \textcolor{blue}{(0.002)} & 0.954 \textcolor{blue}{(0.002)} & 0.935 \textcolor{blue}{(0.005)} \\
        GCN & 0.745 \textcolor{blue}{(0.016)} & 0.778 \textcolor{blue}{(0.011)} & 0.829 \textcolor{blue}{(0.009)} & 0.805 \textcolor{blue}{(0.008)} \\
        GAT & 0.743 \textcolor{blue}{(0.033)} & 0.809 \textcolor{blue}{(0.013)} & 0.851 \textcolor{blue}{(0.029)} & 0.789 \textcolor{blue}{(0.046)} \\
        MLP & 0.805 \textcolor{blue}{(0.013)} & 0.819 \textcolor{blue}{(0.010)} & 0.897 \textcolor{blue}{(0.005)} & 0.872 \textcolor{blue}{(0.007)} \\
        \bottomrule
    \end{tabular}
    }
    \label{tab:test_results}
\end{table}

We conducted a comprehensive evaluation on our evaluation set  across different architectural paradigms. 
We randomly selected 29, 6, and 6 quarters as the training, validation, and test data, respectively, with an approximate ratio of 70:15:15.
Detailed experimental configurations, including hyperparameter settings, training protocols, and hardware specifications, are provided in Appendix \ref{Appendix_Experimental Setup}. 
The comparative results of the model performance are illustrated in Table~\ref{tab:test_results}.
Multiple evaluation metrics were employed to assess our model performance, including Precision, F1 score, Area Under the Receiver Operating Characteristic curve (AUC), and Area Under the Precision-Recall curve (PRAUC).


Compared with GCN and GAT, MLP achieves higher overall performance across all metrics, with a precision of 80.5\%, an F1 score of 81.9\%, an AUC of 89.7\%, and a PRAUC of 87.2\%. This suggests that even a simple feedforward architecture can effectively leverage the higher-order feature representations provided by HODDI, highlighting the dataset's well-structured and informative nature. Among the GNN models, GAT and GCN demonstrate comparable performance. While GAT achieves a higher F1 score (80.9\% vs. 77.8\%) and AUC (85.1\% vs. 82.9\%) compared to GCN, the latter performs better in terms of F1 score and PRAUC. That means that HODDI is well-suited for both the MLP model and GNN models.

By leveraging hypergraph architectures, HyGNN and HGNN-SA, which effectively capture higher-order relationships in drug-side effect interactions, the results can be significantly improved. The HGNN-SA outperforms all other approaches, achieving a precision of 90.6\%, an F1 score of 93.3\%, an AUC of 95.7\%, and a PRAUC of 93.9\%. 
These results demonstrate that HGNN-based models which explicitly integrate hypergraph structures can effectively leverage the higher-order drug-drug interaction relationships captured in our HODDI dataset.


Additionally, the low standard deviations across all models (ranging from 0.1\% to 4.6\%) indicate robust and stable learning behavior, further validating the dataset's reliability for capturing the interaction between drugs and side effects. The strong performance of both GNN models and the MLP model suggests that HODDI provides rich and high-quality feature representations, making it a valuable resource for advancing machine learning approaches in drug-drug interaction studies.

\section{Conclusions}
In this paper, we introduced HODDI, the first comprehensive dataset capturing higher-order drug-drug interactions. Our HODDI dataset comprises 109,744 records with 2,506 unique drugs and 4,569 unique side effects. Through extensive evaluation, we demonstrated that higher-order features significantly enhance prediction performance, with hypergraph-based model outperforming traditional approaches by effectively capturing multi-drug relationships. These findings underscore the importance of higher-order data in advancing pharmacovigilance, drug safety, and personalized medicine. 
Future research opportunities based on HODDI include the development of advanced heterogeneous hypergraph neural networks that leverage attention mechanisms and novel message-passing schemes, the construction of dynamic hypergraph datasets with temporal information, and the integration of large language models with multimodal data. Our work reveals that HODDI establishes a robust foundation for investigating polypharmacy adverse effects, contributing to safer and more effective pharmacological interventions.

\section*{Impact Statement}
This paper introduces a benchmark dataset for higher-order drug-drug interactions, addressing a critical need in healthcare and pharmaceutical research. The dataset enables systematic study of complex drug interactions and their side effects, potentially improving drug safety assessment and patient care. This data resource supports application-driven machine learning research in pharmacology while facilitating development of advanced models for predicting adverse drug reactions.

\bibliography{example_paper}
\bibliographystyle{icml2025}

\clearpage

\appendix
\onecolumn
\section{Appendix}
\subsection{HODDI Dataset Construction: Processing Details}
\label{Appendix_A0-1}
\FloatBarrier

\label{HODDI_construction_algo}
\begin{itemize}
    \item \textbf{Data Collection and Preprocessing}
    \begin{itemize}
        \item Download and extract XML files from FAERS quarterly datasets (2014Q3-2024Q3)
        \item Extract key components from FAERS records: report ID, drug information (standardized drug name and drug role), and side effect
        \item Filter records based on \textit{Drug Role} counts:
           \begin{itemize}
               \item Condition 1: Count (\textit{Drug Role 3}) $\geq$ 2, Count (\textit{Drug Role1}) = 0, Count (\textit{Drug Role 2}) = 0
               \item Condition 2: Count (\textit{Drug Role 1}) $\geq$ 1, Count (\textit{Drug Role 3}) $\geq$ 1, Count (\textit{Drug Role 2}) = 0
               \item Condition 3: Count (\textit{Drug Role 1}) $\geq$ 1, Count (\textit{Drug Role 2}) $\geq$ 1, Count (\textit{Drug Role 3}) $\geq$ 1
           \end{itemize}
    \end{itemize}

    \item \textbf{Standardized Drug Names Processing}
       \begin{itemize}
           \item Normalization of Standardized drug names:
               \begin{itemize}
                   \item Convert to uppercase
                   \item Remove salt form suffixes
                   \item Handle compound names
               \end{itemize}
           \item Map normalized drug names to DrugBank IDs
    \end{itemize}
    
    \item \textbf{Side Effects Processing}
       \begin{itemize}
           \item Generate 768-dimensional embeddings for FAERS side effects and MedDRA terms using SapBERT
           \item Calculate cosine similarities between SapBERT-based embeddings of FAERS side effects and MedDRA terms
           \item Map FAERS side effects to recommended MedDRA terms by highest similarity
           \item Query recommended UMLS CUIs from UMLS Metathesaurus using MedDRA terms
           \item Stratify recommended UMLS CUIs by the cosine similarity threshold (e.g., 0.9)
       \end{itemize}

    \item \textbf{Dataset Construction}
    \begin{itemize}
        \item Generate positive samples:
            \begin{itemize}
                \item Select records based on the predefined cosine similarity threshold
                \item Remove duplicate records and drug combination supersets
            \end{itemize}
        \item Generate negative samples:
            \begin{itemize}
                \item For each positive sample:
                    \begin{itemize}
                        \item Randomly replace one drug and one side effect from their respective complement sets
                        \item Verify that the generated negative sample is absent in all positive samples
                    \end{itemize}
            \end{itemize}
        \item Create evaluation subsets for benchmark model evaluation:
           \begin{itemize}
               \item Evaluation Subset 1: A set of records where each record contains 2-8 drugs and their associated side effects occur with a frequency of 5-50 times across all records
               \item Evaluation Subset 2: A set of records where each record contains 2-16 drugs and their associated side effects occur with a frequency of 5-50 times across all records
               \item Evaluation Subset 3: A set of records where each record contains 2-16 drugs and their associated side effects occur with a frequency of 5-100 times across all records
           \end{itemize}
    \end{itemize}
\end{itemize}
\clearpage


\clearpage
\subsection{HODDI Dataset Structure}
\label{Appendix_A0-2}

\vspace{2mm}

Table~\ref{hgnn_dataset_structure} outlines the evaluation dataset structure, which intrinsically captures the higher-order drug-drug interactions relationships. Each record contains a FAERS report identifier, a recommended UMLS CUI code for adverse effect (with cosine similarity $\geq$ 0.9), and a list of DrugBank identifiers representing co-administered drugs. The hyperedge label (-1/1) indicates whether the drug combination leads to the specified side effect. Additional metadata includes the record's temporal information and the number of drugs involved in each record.

\vspace{2mm}

Table~\ref{gnn_dataset_structure} outlines the converted evaluation dataset structure, which examines the pairwise drug interactions. Each record consists of a FAERS report identifier, a source and target drug pair represented by their DrugBank IDs, and a recommended UMLS CUI representing the observed side effect caused by the drug pair. The edge label (-1/1) denotes whether this drug pair causes the specified adverse effect.

\vspace{2mm}

\renewcommand{\arraystretch}{1.2} 
\begin{table}[!ht]
   \centering
   \small 
   \caption{Data structure of the evaluation set from HODDI dataset.}
   \resizebox{0.95\textwidth}{!}{
   \begin{tabular}{@{}lll@{}}
       \toprule
       \rowcolor{gray!20}
       \multicolumn{1}{c}{\textbf{Variable}} & \multicolumn{1}{c}{\textbf{Column Name}} & \multicolumn{1}{c}{\textbf{Description}} \\ 
       \midrule
       Report ID & report\_id & FAERS report identifier (negative samples end with "n") \\
       Recommend UMLS CUI & SE\_above\_0.9 & High-confidence UMLS CUIs identified using a cosine similarity threshold of 0.9 \\
       DrugBank ID & DrugBankID & List of standardized DrugBank identifiers \\
       Hyperedge Label & hyperedge\_label & Binary label for hyperedge existence (1: positive samples; -1: negative samples) \\
       Record Time & time & Year and quarter of the record (e.g., 2014Q3) \\
       Drug Count & DrugBankID\_list\_length & Number of co-administered drugs \\
       Row Index & row\_index & Row index in merged dataset (2014Q3-2024Q3) \\
       \bottomrule
   \end{tabular}}
   \label{hgnn_dataset_structure}
\end{table}
\renewcommand{\arraystretch}{1} 
\vspace{2mm}

\renewcommand{\arraystretch}{1.2} 
\begin{table}[!ht]
   \centering
   \small 
   \caption{Data structure of the converted evaluation set from HODDI dataset.}
   \begin{tabularx}{0.95\textwidth}{@{}p{2.6cm}p{2.6cm}X@{}} 
       \toprule
       \rowcolor{gray!20}
       \multicolumn{1}{c}{\textbf{\footnotesize Variable}} & \multicolumn{1}{c}{\textbf{\footnotesize Column Name}} & \multicolumn{1}{c}{\textbf{\footnotesize Description}} \\ 
       \midrule
       Report ID & report\_id & FAERS report identifier (negative samples end with "n") \\
       Source Drug & source & Drug node in undirected graph \\
       Target Drug & target & Drug node in undirected graph \\
       Side Effect CUI & SE\_label & High-confidence UMLS CUIs identified using a cosine similarity threshold of 0.9 \\
       Edge Label & edge\_label & Binary label for edge existence (1: positive samples; -1: negative samples) \\
       \bottomrule
   \end{tabularx}
   \label{gnn_dataset_structure}
\end{table}
\renewcommand{\arraystretch}{1}

\clearpage
\subsection{HODDI Dataset Statistics}
\label{Appendix_A1}

In the HODDI dataset spanning from 2014Q3 to 2024Q3, we analyzed the distribution patterns of drug combination sizes and occurrence frequency of side effects.
As shown in Table~\ref{drugs_per_record}, 80.48\% of records contain between 2-10 drugs, with 54,304 (48.89\%) records containing 2-5 drugs and 35,087 (31.59\%) records containing 6-10 drugs.
The frequency decreases significantly for records with higher drug counts, with only 1,234 records (1.11\% of the total) containing no less than 30 drugs.

Table~\ref{se_positive} presents the distribution of side effect occurrences in positive samples from 2014Q3 to 2024Q3. The data shows a highly left-skewed and long-tailed distribution where 2,501 side effects (63.32\%) appear only 1-10 times, while only 3 side effects occur more than 1,000 times. The majority of side effects (82.63\%) occur less than 30 times, indicating that most side effects are relatively uncommon in the HODDI dataset.

Table~\ref{se_negative} presents the distribution of side effect occurrences in negative samples from 2014Q3 to 2024Q3. After resampling, the data exhibits a near-Gaussian distribution with a slight left skew, where 3,146 side effects (68.69\%) occur 21-30 times. The distribution is more concentrated compared to positive samples, with no occurrences below 10 times and few (15 cases, 0.33\%) above 40 times.

\begin{table}[h]
   \centering
   \caption{Distribution of drug count per record in positive and negative samples (2014Q3-2024Q3). (\#Drug/Record: drug count per record; \#Record: number of records.)}
   \resizebox{0.25\textwidth}{!}{ 
   \begin{tabular}{@{}>{\centering\arraybackslash}p{3cm}>{\centering\arraybackslash}p{2cm}@{}}
       \toprule
       \rowcolor{gray!20}
       \textbf{\large \#Drug/Record} & \textbf{\large \#Record} \\ 
       \midrule
       1 & 474 \\
       2-5 & 54,304 \\
       6-10 & 35,087 \\
       11-15 & 12,730 \\
       16-20 & 4,661 \\
       21-30 & 2,582 \\
       31-40 & 798 \\
       41-50 & 176 \\
       51-100 & 260 \\
       101+ & 0 \\
       \bottomrule
   \end{tabular}}
   
   \label{drugs_per_record}
\end{table}

\begin{table}[!h]
   \centering
   \caption{Distribution of side effect occurrences in positive samples (2014Q3-2024Q3). (\#Occurrence: number of side effect occurrences in positive samples; \#Side Effect: number of side effects in positive samples.)}
   \resizebox{0.25\textwidth}{!}{ 
   \begin{tabular}{@{}>{\centering\arraybackslash}p{3cm}>{\centering\arraybackslash}p{3cm}@{}}
       \toprule
       \rowcolor{gray!20}
       \textbf{\large  \#Occurrence} & \textbf{\large 
 \#Side Effect} \\ 
       \midrule
       1-10 & 2,501 \\
       11-20 & 535 \\
       21-30 & 228 \\
       31-40 & 126 \\
       41-50 & 91 \\
       51-100 & 221 \\
       101-200 & 137 \\
       201-500 & 85 \\
       501-1000 & 23 \\
       1001-5000 & 3 \\
       \bottomrule
   \end{tabular}}
   
   \label{se_positive}
\end{table}

\begin{table}[!h]
   \centering
   \caption{Distribution of side effect occurrences in negative samples (2014Q3-2024Q3). (\#Occurrence: number of side effect occurrences in negative samples; \#Side Effect: number of side effects in negative samples.)}
   \resizebox{0.25\textwidth}{!}{ 
   \begin{tabular}{@{}>{\centering\arraybackslash}p{3cm}>{\centering\arraybackslash}p{3cm}@{}}
       \toprule
       \rowcolor{gray!20}
       \textbf{\large \#Occurrence} & \textbf{\large \#Side Effect} \\ 
       \midrule
       1-10 & 0 \\
       11-20 & 773 \\
       21-30 & 3,146 \\
       31-40 & 646 \\
       41-50 & 15 \\
       \bottomrule
   \end{tabular}}
   
   \label{se_negative}
\end{table}

\clearpage
\begin{table*}[!ht]
    \centering
    \caption{Quarterly distribution of drug conditions and cosine similarity ranges in HODDI dataset (2014Q3-2024Q3). Black numbers show record counts for each drug condition and cosine similarity range, while blue numbers in parentheses indicate their proportions of total records. Total, Mean, and Std Dev rows present aggregate statistics for each column.}
    \resizebox{0.9\textwidth}{!}{
    \renewcommand{\arraystretch}{1.2}
    \begin{tabular}{@{}l|ccc|c|ccc|c@{}}
        \toprule
        \rowcolor{gray!20}
        & \multicolumn{4}{c|}{\textbf{\large Drug Condition}} & \multicolumn{4}{c}{\textbf{\large Cosine similarity}} \\
        \cmidrule(r){2-5} \cmidrule(l){6-9}
        \rowcolor{gray!20}
        \textbf{\large Time} & \textbf{1} & \textbf{2} & \textbf{3} & \textbf{Total} & \textbf{< 0.8} & \textbf{0.8 - 0.9} & \textbf{> 0.9} & \textbf{Total} \\
        \midrule
        \rowcolor{white}
        \textbf{2014Q3}  & 33 \textcolor{blue}{(0.02)} & 598 \textcolor{blue}{(0.44)} & 719 \textcolor{blue}{(0.53)} & 1350 & 193 \textcolor{blue}{(0.03)} & 171 \textcolor{blue}{(0.03)} & 5906 \textcolor{blue}{(0.94)} & 6270 \\
        \rowcolor{gray!8}
        \textbf{2014Q4}  & 27 \textcolor{blue}{(0.02)} & 515 \textcolor{blue}{(0.44)} & 641 \textcolor{blue}{(0.54)} & 1183 & 214 \textcolor{blue}{(0.04)} & 129 \textcolor{blue}{(0.02)} & 4847 \textcolor{blue}{(0.93)} & 5190 \\
        \rowcolor{white}
        \textbf{2015Q1}  & 30 \textcolor{blue}{(0.02)} & 561 \textcolor{blue}{(0.43)} & 718 \textcolor{blue}{(0.55)} & 1309 & 200 \textcolor{blue}{(0.03)} & 174 \textcolor{blue}{(0.03)} & 5526 \textcolor{blue}{(0.94)} & 5900 \\
        \rowcolor{gray!8}
        \textbf{2015Q2}  & 32 \textcolor{blue}{(0.02)} & 778 \textcolor{blue}{(0.45)} & 936 \textcolor{blue}{(0.54)} & 1746 & 308 \textcolor{blue}{(0.04)} & 235 \textcolor{blue}{(0.03)} & 7822 \textcolor{blue}{(0.94)} & 8365 \\
        \rowcolor{white}
        \textbf{2015Q3}  & 49 \textcolor{blue}{(0.03)} & 773 \textcolor{blue}{(0.40)} & 1092 \textcolor{blue}{(0.57)} & 1914 & 281 \textcolor{blue}{(0.03)} & 317 \textcolor{blue}{(0.04)} & 8196 \textcolor{blue}{(0.93)} & 8794 \\
        \rowcolor{gray!8}
        \textbf{2015Q4}  & 33 \textcolor{blue}{(0.02)} & 711 \textcolor{blue}{(0.38)} & 1138 \textcolor{blue}{(0.60)} & 1882 & 318 \textcolor{blue}{(0.04)} & 244 \textcolor{blue}{(0.03)} & 7963 \textcolor{blue}{(0.93)} & 8525 \\
        \rowcolor{white}
        \textbf{2016Q1}  & 39 \textcolor{blue}{(0.02)} & 717 \textcolor{blue}{(0.38)} & 1122 \textcolor{blue}{(0.60)} & 1878 & 385 \textcolor{blue}{(0.04)} & 269 \textcolor{blue}{(0.03)} & 7960 \textcolor{blue}{(0.92)} & 8614 \\
        \rowcolor{gray!8}
        \textbf{2016Q2}  & 55 \textcolor{blue}{(0.03)} & 869 \textcolor{blue}{(0.43)} & 1120 \textcolor{blue}{(0.55)} & 2044 & 437 \textcolor{blue}{(0.05)} & 300 \textcolor{blue}{(0.03)} & 8695 \textcolor{blue}{(0.92)} & 9432 \\
        \rowcolor{white}
        \textbf{2016Q3}  & 53 \textcolor{blue}{(0.02)} & 1154 \textcolor{blue}{(0.49)} & 1132 \textcolor{blue}{(0.48)} & 2339 & 432 \textcolor{blue}{(0.04)} & 310 \textcolor{blue}{(0.03)} & 9363 \textcolor{blue}{(0.93)} & 10105 \\
        \rowcolor{gray!8}
        \textbf{2016Q4}  & 40 \textcolor{blue}{(0.02)} & 830 \textcolor{blue}{(0.47)} & 879 \textcolor{blue}{(0.50)} & 1749 & 401 \textcolor{blue}{(0.05)} & 255 \textcolor{blue}{(0.03)} & 7503 \textcolor{blue}{(0.92)} & 8159 \\
        \rowcolor{white}
        \textbf{2017Q1}  & 44 \textcolor{blue}{(0.02)} & 821 \textcolor{blue}{(0.45)} & 953 \textcolor{blue}{(0.52)} & 1818 & 372 \textcolor{blue}{(0.04)} & 282 \textcolor{blue}{(0.03)} & 7642 \textcolor{blue}{(0.92)} & 8296 \\
        \rowcolor{gray!8}
        \textbf{2017Q2}  & 39 \textcolor{blue}{(0.02)} & 900 \textcolor{blue}{(0.48)} & 937 \textcolor{blue}{(0.50)} & 1876 & 404 \textcolor{blue}{(0.05)} & 299 \textcolor{blue}{(0.04)} & 7490 \textcolor{blue}{(0.91)} & 8193 \\
        \rowcolor{white}
        \textbf{2017Q3}  & 41 \textcolor{blue}{(0.02)} & 855 \textcolor{blue}{(0.44)} & 1042 \textcolor{blue}{(0.54)} & 1938 & 478 \textcolor{blue}{(0.05)} & 301 \textcolor{blue}{(0.03)} & 8659 \textcolor{blue}{(0.92)} & 9438 \\
        \rowcolor{gray!8}
        \textbf{2018Q1}  & 60 \textcolor{blue}{(0.02)} & 1280 \textcolor{blue}{(0.47)} & 1386 \textcolor{blue}{(0.51)} & 2726 & 599 \textcolor{blue}{(0.05)} & 516 \textcolor{blue}{(0.04)} & 11285 \textcolor{blue}{(0.91)} & 12400 \\
        \rowcolor{white}
        \textbf{2018Q2}  & 54 \textcolor{blue}{(0.02)} & 1643 \textcolor{blue}{(0.48)} & 1691 \textcolor{blue}{(0.50)} & 3388 & 855 \textcolor{blue}{(0.05)} & 733 \textcolor{blue}{(0.04)} & 16909 \textcolor{blue}{(0.91)} & 18497 \\
        \rowcolor{gray!8}
        \textbf{2018Q3}  & 57 \textcolor{blue}{(0.02)} & 1744 \textcolor{blue}{(0.51)} & 1641 \textcolor{blue}{(0.48)} & 3442 & 924 \textcolor{blue}{(0.05)} & 636 \textcolor{blue}{(0.04)} & 16286 \textcolor{blue}{(0.91)} & 17846 \\
        \rowcolor{white}
        \textbf{2018Q4}  & 76 \textcolor{blue}{(0.02)} & 1557 \textcolor{blue}{(0.47)} & 1654 \textcolor{blue}{(0.50)} & 3287 & 814 \textcolor{blue}{(0.05)} & 641 \textcolor{blue}{(0.04)} & 14740 \textcolor{blue}{(0.91)} & 16195 \\
        \rowcolor{gray!8}
        \textbf{2019Q1}  & 67 \textcolor{blue}{(0.02)} & 1609 \textcolor{blue}{(0.47)} & 1745 \textcolor{blue}{(0.51)} & 3421 & 845 \textcolor{blue}{(0.05)} & 731 \textcolor{blue}{(0.04)} & 15448 \textcolor{blue}{(0.91)} & 17024 \\
        \rowcolor{white}
        \textbf{2019Q2}  & 60 \textcolor{blue}{(0.02)} & 1524 \textcolor{blue}{(0.47)} & 1681 \textcolor{blue}{(0.51)} & 3265 & 878 \textcolor{blue}{(0.05)} & 679 \textcolor{blue}{(0.04)} & 14651 \textcolor{blue}{(0.90)} & 16208 \\
        \rowcolor{gray!8}
        \textbf{2019Q3}  & 71 \textcolor{blue}{(0.02)} & 1726 \textcolor{blue}{(0.49)} & 1749 \textcolor{blue}{(0.49)} & 3546 & 901 \textcolor{blue}{(0.05)} & 699 \textcolor{blue}{(0.04)} & 15704 \textcolor{blue}{(0.91)} & 17304 \\
        \rowcolor{white}
        \textbf{2019Q4}  & 72 \textcolor{blue}{(0.02)} & 1473 \textcolor{blue}{(0.46)} & 1639 \textcolor{blue}{(0.51)} & 3184 & 877 \textcolor{blue}{(0.06)} & 612 \textcolor{blue}{(0.04)} & 14311 \textcolor{blue}{(0.91)} & 15800 \\
        \rowcolor{gray!8}
        \textbf{2020Q1}  & 62 \textcolor{blue}{(0.02)} & 1586 \textcolor{blue}{(0.44)} & 1935 \textcolor{blue}{(0.54)} & 3583 & 897 \textcolor{blue}{(0.05)} & 609 \textcolor{blue}{(0.04)} & 15719 \textcolor{blue}{(0.91)} & 17225 \\
        \rowcolor{white}
        \textbf{2020Q2}  & 59 \textcolor{blue}{(0.02)} & 1570 \textcolor{blue}{(0.46)} & 1810 \textcolor{blue}{(0.53)} & 3439 & 1014 \textcolor{blue}{(0.06)} & 617 \textcolor{blue}{(0.04)} & 14650 \textcolor{blue}{(0.90)} & 16281 \\
        \rowcolor{gray!8}
        \textbf{2020Q3}  & 81 \textcolor{blue}{(0.02)} & 1630 \textcolor{blue}{(0.47)} & 1730 \textcolor{blue}{(0.50)} & 3441 & 1417 \textcolor{blue}{(0.08)} & 772 \textcolor{blue}{(0.04)} & 16243 \textcolor{blue}{(0.88)} & 18432 \\
        \rowcolor{white}
        \textbf{2020Q4}  & 59 \textcolor{blue}{(0.02)} & 1393 \textcolor{blue}{(0.50)} & 1334 \textcolor{blue}{(0.48)} & 2786 & 862 \textcolor{blue}{(0.06)} & 521 \textcolor{blue}{(0.04)} & 12734 \textcolor{blue}{(0.90)} & 14117 \\
        \rowcolor{gray!8}
        \textbf{2021Q1}  & 72 \textcolor{blue}{(0.03)} & 1407 \textcolor{blue}{(0.53)} & 1185 \textcolor{blue}{(0.44)} & 2664 & 933 \textcolor{blue}{(0.07)} & 563 \textcolor{blue}{(0.04)} & 12562 \textcolor{blue}{(0.89)} & 14058 \\
        \rowcolor{white}
        \textbf{2021Q2}  & 82 \textcolor{blue}{(0.03)} & 1455 \textcolor{blue}{(0.52)} & 1271 \textcolor{blue}{(0.45)} & 2808 & 1078 \textcolor{blue}{(0.07)} & 554 \textcolor{blue}{(0.04)} & 13083 \textcolor{blue}{(0.89)} & 14715 \\
        \rowcolor{gray!8}
        \textbf{2021Q3}  & 63 \textcolor{blue}{(0.02)} & 1289 \textcolor{blue}{(0.49)} & 1270 \textcolor{blue}{(0.48)} & 2622 & 1178 \textcolor{blue}{(0.08)} & 653 \textcolor{blue}{(0.04)} & 13341 \textcolor{blue}{(0.88)} & 15172 \\
        \rowcolor{white}
        \textbf{2021Q4}  & 801 \textcolor{blue}{(0.53)} & 322 \textcolor{blue}{(0.21)} & 377 \textcolor{blue}{(0.25)} & 1500 & 640 \textcolor{blue}{(0.07)} & 381 \textcolor{blue}{(0.04)} & 7756 \textcolor{blue}{(0.88)} & 8777 \\
        \rowcolor{gray!8}
        \textbf{2022Q1}  & 856 \textcolor{blue}{(0.51)} & 362 \textcolor{blue}{(0.22)} & 453 \textcolor{blue}{(0.27)} & 1671 & 852 \textcolor{blue}{(0.08)} & 555 \textcolor{blue}{(0.05)} & 9595 \textcolor{blue}{(0.87)} & 11002 \\
        \rowcolor{white}
        \textbf{2022Q2}  & 906 \textcolor{blue}{(0.59)} & 262 \textcolor{blue}{(0.17)} & 377 \textcolor{blue}{(0.24)} & 1545 & 828 \textcolor{blue}{(0.08)} & 494 \textcolor{blue}{(0.05)} & 8810 \textcolor{blue}{(0.87)} & 10132 \\
        \rowcolor{gray!8}
        \textbf{2022Q3}  & 906 \textcolor{blue}{(0.54)} & 328 \textcolor{blue}{(0.20)} & 443 \textcolor{blue}{(0.26)} & 1677 & 837 \textcolor{blue}{(0.08)} & 442 \textcolor{blue}{(0.04)} & 9014 \textcolor{blue}{(0.88)} & 10293 \\
        \rowcolor{white}
        \textbf{2022Q4}  & 849 \textcolor{blue}{(0.52)} & 287 \textcolor{blue}{(0.17)} & 512 \textcolor{blue}{(0.31)} & 1648 & 850 \textcolor{blue}{(0.08)} & 451 \textcolor{blue}{(0.04)} & 9537 \textcolor{blue}{(0.88)} & 10838 \\
        \rowcolor{gray!8}
        \textbf{2023Q1}  & 1007 \textcolor{blue}{(0.59)} & 333 \textcolor{blue}{(0.19)} & 374 \textcolor{blue}{(0.22)} & 1714 & 762 \textcolor{blue}{(0.08)} & 469 \textcolor{blue}{(0.05)} & 8468 \textcolor{blue}{(0.87)} & 9699 \\
        \rowcolor{white}
        \textbf{2023Q2}  & 1031 \textcolor{blue}{(0.58)} & 350 \textcolor{blue}{(0.20)} & 386 \textcolor{blue}{(0.22)} & 1767 & 770 \textcolor{blue}{(0.08)} & 398 \textcolor{blue}{(0.04)} & 8540 \textcolor{blue}{(0.88)} & 9708 \\
        \rowcolor{gray!8}
        \textbf{2023Q3}  & 954 \textcolor{blue}{(0.58)} & 354 \textcolor{blue}{(0.22)} & 331 \textcolor{blue}{(0.20)} & 1639 & 729 \textcolor{blue}{(0.08)} & 359 \textcolor{blue}{(0.04)} & 7857 \textcolor{blue}{(0.88)} & 8945 \\
        \rowcolor{white}
        \textbf{2023Q4}  & 900 \textcolor{blue}{(0.55)} & 365 \textcolor{blue}{(0.22)} & 361 \textcolor{blue}{(0.22)} & 1626 & 834 \textcolor{blue}{(0.08)} & 388 \textcolor{blue}{(0.04)} & 8631 \textcolor{blue}{(0.88)} & 9853 \\
        \rowcolor{gray!8}
        \textbf{2024Q1}  & 820 \textcolor{blue}{(0.51)} & 353 \textcolor{blue}{(0.22)} & 425 \textcolor{blue}{(0.27)} & 1598 & 787 \textcolor{blue}{(0.09)} & 406 \textcolor{blue}{(0.04)} & 7872 \textcolor{blue}{(0.87)} & 9065 \\
        \rowcolor{white}
        \textbf{2024Q2}  & 780 \textcolor{blue}{(0.51)} & 348 \textcolor{blue}{(0.23)} & 406 \textcolor{blue}{(0.26)} & 1534 & 801 \textcolor{blue}{(0.10)} & 360 \textcolor{blue}{(0.04)} & 7051 \textcolor{blue}{(0.86)} & 8212 \\
        \rowcolor{gray!8}
        \textbf{2024Q3}  & 821 \textcolor{blue}{(0.54)} & 326 \textcolor{blue}{(0.22)} & 362 \textcolor{blue}{(0.24)} & 1509 & 710 \textcolor{blue}{(0.09)} & 334 \textcolor{blue}{(0.04)} & 6992 \textcolor{blue}{(0.87)} & 8036 \\
        \midrule
        \rowcolor{gray!20}
        \textbf{Total} & 12172 \textcolor{blue}{(0.13)} & 37861 \textcolor{blue}{(0.41)} & 42031 \textcolor{blue}{(0.46)} & 92064 & 28331 \textcolor{blue}{(0.06)} & 18170 \textcolor{blue}{(0.04)} & 423643 \textcolor{blue}{(0.90)} & 470144 \\
        \rowcolor{gray!20}
        \textbf{Mean} & 296.88 \textcolor{blue}{(0.13)} & 923.44 \textcolor{blue}{(0.41)} & 1025.15 \textcolor{blue}{(0.46)} & 2246 & 6910 \textcolor{blue}{(0.39)} & 443.17 \textcolor{blue}{(0.03)} & 10332.76 \textcolor{blue}{(0.58)} & 17686 \\
        \rowcolor{gray!20}
        \textbf{Std Dev} & 386.15 \textcolor{blue}{(0.27)} & 513.34 \textcolor{blue}{(0.36)} & 512.72 \textcolor{blue}{(0.36)} & 1412 & 287.83 \textcolor{blue}{(0.05)} & 1752 \textcolor{blue}{(0.32)} & 3474.49 \textcolor{blue}{(0.63)} & 5514 \\
        \bottomrule
    \end{tabular}}
    \label{combined_stats_all_quarters}
\end{table*}

\clearpage
\subsection{Experimental Setup}
\label{Appendix_Experimental Setup}

\subsubsection{Training Details}

\paragraph{MLP.}
The drug and side effect representations are each projected from 768 to 64 dimensions through separate reduction layers. The resulting embeddings are concatenated and passed through a classifier with hidden layers of 128 and 32 dimensions, followed by a final output layer with a single neuron, using ReLU activation. The model is trained using the Adam optimizer with a learning rate of 0.001 and Binary Cross Entropy Loss. Training runs for a maximum of 500 epochs, with early stopping applied if the validation loss does not improve by at least 0.001 for 20 consecutive epochs. 

\paragraph{GCN and GAT.}
The GCN and GAT models share the same architecture, with the only difference being the use of GCN layers in the GCN model and GAT layers in the GAT model. The architecture consists of an input layer, followed by a hidden layer with 128 channels, a second layer reducing the dimensionality to 64 channels, and a final classifier layer with 2 output neurons. Both models are trained with a learning rate of 0.0005, weight decay of 0.001, and use AdamW as the optimizer. The loss function is Cross Entropy Loss with class weights, and early stopping is applied if the validation loss does not improve by 0.001 for 20 consecutive epochs. A learning rate scheduler, ReduceLROnPlateau, reduces the learning rate by a factor of 0.1 if the validation loss plateaus for 20 epochs. The training data is split into 29 quarters for training, 6 quarters for validation, and 6 quarters for testing, with a loss weight balance of 1.0, ensuring equal weighting between drug-drug and drug-side effect losses. Training is conducted for a maximum of 500 epochs.

\paragraph{HGNN-Based Method.} We introduce HGNN with SMILES Embedding Features and Attention Mechanism, a novel architecture specifically crafted to capture high-order relationships in chemical data efficiently. By combining hypergraph convolution, multi-head attention mechanisms, and molecular embeddings derived from SMILES representations, our approach significantly enhances feature representation and boosts predictive accuracy. Unlike conventional graph neural networks (GNNs), this model utilizes a hypergraph incidence matrix to model intricate interactions, where a single hyperedge can link multiple drug entities. Node features are derived from SMILES (Simplified Molecular Input Line Entry System) representations and converted into numerical embeddings via a SMILES-to-vector algorithm. These features undergo linear transformation and feature fusion before being processed through hypergraph convolution layers. To improve the model's expressive power, a multi-head attention mechanism is integrated into the hypergraph convolution process, enabling the dynamic assignment of importance to different hyperedges and facilitating more adaptable feature representations. The model architecture comprises an initial feature encoder, a series of hypergraph convolution layers (HypergraphConv), and a fully connected output layer for classification. The number of hypergraph convolution layers, denoted as $L$, is adjustable, allowing the model to capture both local and global dependencies in hypergraph-structured drug-side effect datasets. During forward propagation, drug node features are first encoded through a ReLU-activated transformation and combined with external SMILES-based embeddings. The hypergraph convolution operation is applied iteratively across $L$ layers, progressively refining node embeddings based on hypergraph connectivity. Finally, hyperedge features are computed by aggregating node representations, followed by a softmax layer for binary classification.

\subsubsection{Hardware Configuration}
\begin{itemize}
    \item Computing Device: CUDA-enabled GPU (when available) or CPU
    \item Memory Management:
    \begin{itemize}
        \item SMILES processing batch size: 16
        \item Maximum sequence length: 256 for SMILES strings
    \end{itemize}
    \item Random Seeds: Fixed seed values of 42, 3407, 54321, and 123456 were used for reproducibility
\end{itemize}

\end{document}